\documentclass[11pt,letterpaper]{article}
\usepackage{emnlp2016}
\usepackage{times}
\usepackage{latexsym}

\usepackage{algorithm}
\usepackage{algorithmic}
\usepackage{amsmath}
\usepackage{amsthm}
\usepackage{amssymb}
\usepackage{pifont}
\usepackage{graphicx}
\usepackage{xspace}
\usepackage{tabularx}
\usepackage{multicol}
\usepackage{multirow}
\usepackage{url}
\usepackage{wrapfig}
\usepackage{comment}
\usepackage{listings}
\usepackage[utf8]{inputenc}
\usepackage{fancyvrb}
\usepackage{booktabs}
\usepackage{color,soul}
\usepackage[normalem]{ulem}

\newcommand{\bmx}[0]{\begin{bmatrix}}
\newcommand{\emx}[0]{\end{bmatrix}}

\newcommand{\vect}[1]{\mathbf{#1}}

\newcommand{\matr}[1]{\mathbf{#1}}

\newcommand{\vb}[0]{\vect{b}}
\newcommand{\vc}[0]{\vect{c}}

\newcommand{\vh}[0]{\vect{h}}

\newcommand{\vz}[0]{\vect{z}}

\newcommand{\mW}[0]{\matr{W}}

\newcommand{\mE}[0]{\matr{E}}

\newcommand{\mU}[0]{\matr{U}}

\newcommand{\init}{\ensuremath{\text{init}}}

\emnlpfinalcopy

\def\emnlppaperid{***}

\title{Zero-Resource Translation with \\ Multi-Lingual Neural Machine Translation}

\author{Orhan Firat$^\star$\\
  Middle East Technical University \\
  {\tt orhan.firat@ceng.metu.edu.tr}
    \\\And
  Baskaran Sankaran \\
  IBM T.J. Watson Research Center \\ \\\AND
  Yaser Al-Onaizan\\
  IBM T.J. Watson Research Center \\ \\\And
  ~~~~~~~~~~~~~~~Fatos T. Yarman Vural\\
  ~~~~~~~~~~~~~~~Middle East Technical University \\ \\\And
  ~~~~~~~~~~~~Kyunghyun Cho\\
  ~~~~~~~~~~~~New York University\\ \\ \\}

\date{}

\begin{document}

\maketitle

\let\thefootnote\relax\footnote{$^\star$ Work carried out while the author was at IBM Research.}

\begin{abstract}
    In this paper, we propose a novel finetuning algorithm for the recently
    introduced multi-way, mulitlingual neural machine translate that enables
    zero-resource machine translation. When used together with novel many-to-one
    translation strategies, we empirically show that this finetuning algorithm
    allows the multi-way, multilingual model to translate a zero-resource
    language pair (1) as well as a single-pair neural translation model trained
    with up to 1M direct parallel sentences of the same language pair and (2)
    better than pivot-based translation strategy, while keeping only one
    additional copy of attention-related parameters.
\end{abstract}

\section{Introduction}

A recently introduced neural machine
translation~\cite{forcada1997recursive,kalchbrenner2013recurrent,sutskever2014sequence,cho2014learning}
has proven to be a platform for new opportunities in machine translation
research. Rather than word-level translation with language-specific
preprocessing, neural machine translation has found to work well with
statistically segmented subword sequences as well as sequences of
characters~\cite{chung2016character,luong2016achieving,sennrich2015neural,ling2015character}.
Also, recent works show that neural machine translation provides a seamless way
to incorporate multiple modalities other than natural language text in
translation~\cite{luong2015multi,caglayan2016does}. Furthermore, neural machine
translation has been found to translate between multiple languages, 
achieving better translation quality by exploiting positive language
transfer~\cite{dong2015multi,firat2016multi,zoph2016multi}.

In this paper, we conduct in-depth investigation into the recently proposed
multi-way, multilingual neural machine translation~\cite{firat2016multi}.
Specifically, we are interested in its potential for zero-resource machine
translation, in which there does not exist any direct parallel examples between
a target language pair. Zero-resource translation has been addressed by
pivot-based translation in traditional machine translation
research~\cite{wu2007pivot,utiyama2007comparison}, but we explore a way to use
the multi-way, multilingual neural model to translate directly from a source to
target language.

In doing so, we begin by studying different translation strategies available in
the multi-way, multilingual model in
Sec.~\ref{sec:strategies}--\ref{sec:strategies-exp}. The strategies include a
usual one-to-one translation as well as variants of many-to-one translation for
multi-source translation~\cite{zoph2016multi}. We empirically show that the
many-to-one strategies significantly outperform the one-to-one strategy.  

We move on to zero-resource translation by first evaluating a vanilla multi-way,
multilingual model on a zero-resource language pair, which revealed that the
vanilla model cannot do zero-resource translation in
Sec.~\ref{sec:no-finetuning}. Based on the many-to-one strategies we proposed
earlier, we design a novel finetuning strategy that does not require any direct
parallel corpus between a target, zero-resource language pair in
Sec.~\ref{sec:finetuning}, which uses the idea of generating a pseudo-parallel
corpus~\cite{sennrich2015improving}. This strategy makes an additional copy of
the attention mechanism and finetunes only this small set of parameters.

Large-scale experiments with Spanish, French and English show that the proposed
finetuning strategy allows the multi-way, multilingual neural translation model
to perform zero-resource translation as well as a single-pair neural translation
model trained with up to 1M true parallel sentences. This result re-confirms the
potential of the multi-way, multilingual model for low/zero-resource
language translation, which was earlier argued by \newcite{firat2016multi}.

\section{Multi-Way, Multilingual\\~~~~~~~Neural Machine Translation}

Recently \newcite{firat2016multi} proposed an extension of attention-based
neural machine translation~\cite{bahdanau2014neural} that can handle multi-way,
multilingual translation with a shared attention mechanism. This model was
designed to handle multiple source and target languages. In this section, we
briefly overview this multi-way, multilingual model. For more detailed
exposition, we refer the reader to \cite{firat2016multi}.

\subsection{Model Description}

The goal of multi-way, multilingual model is to build a neural translation model
that can translate a source sentence given in one of $N$ languages into one of
$M$ target languages. Thus to handle those $N$ source and $M$ target languages,
the model consists of $N$ encoders and $M$ decoders. Unlike these
language-specific encoders and decoders, only a single attention mechanism is
shared across all $M \times N$ language pairs.

\paragraph{Encoder}
An encoder for the $n$-th source language reads a source sentence $X=(x_1,
\ldots, x_{T_x})$ as a sequence of linguistic symbols and returns a set of
context vectors $C^n=\left\{ \vh_1^n, \ldots, \vh_{T_x}^n \right\}$. The encoder is
usually implemented as a bidirectional recurrent network~\cite{Schuster1997}, and
each context vector $\vh_t^n$ is a concatenation of the forward and reverse
recurrent networks' hidden states at time $t$. Without loss of generality, we
assume that the dimensionalities of the context vector for all source languages
are all same.

\paragraph{Decoder and Attention Mechanism}
A decoder for the $m$-th target language is a conditional recurrent language
model~\cite{mikolov2010recurrent}. At each time step $t'$, it updates its hidden
state by
\[
    \vz_{t'}^m = \varphi^m(\vz_{t'-1}^m, \tilde{y}_{t'-1}^m, \vc_{t'}^m),
\]
based on the previous hidden state $\vz_{t'-1}^m$, previous target symbol
$\tilde{y}_{t'-1}^m$ and the time-dependent context vector $\vc_{t'}^m$.
$\varphi^m$ is a gated recurrent unit (GRU, \cite{cho2014learning}).

The time-dependent context vector is computed by the shared attention mechanism
as a weighted sum of the context vectors from the encoder $C^n$: 
\begin{align}
    \label{eq:cv}
    \vc_{t'}^m = \mU \sum_{t=1}^{T_x} \alpha_{t,t'}^{m,n} \vh_t^n + \vb,
\end{align}
where 
\begin{align}
    \label{eq:att}
    \alpha_{t,t'}^{m,n} \propto \exp\left(
        f_{\text{score}}(\mW^n \vh_t^n, \mW^m \vz_{t'-1}^m, \tilde{y}_{t'-1}^m)
    \right).
\end{align}
The scoring function $f_{\text{score}}$ returns a scalar and is implemented as a
feedforward neural network with a single hidden layer. For more variants of the
attention mechanism for machine translation, see \cite{luong2015effective}.

The initial hidden state of the decoder is initialized as
\begin{align}
    \label{eq:init}
    \vz_{0}^m = \phi_{\text{init}}^m(\mW^n \vh_t^n).
\end{align}

With the new hidden state $\vz_{t'}^m$, the probability distribution over the
next symbol is computed by
\begin{align}
    \label{eq:output}
    p(y_t = w|\tilde{y}_{< t}, X^n) \propto
    \exp(g^m_w(\vz_t^m, \vc_t^m, \mE^m_y\left[ \tilde{y}_{t-1} \right]),
\end{align}
where $g^m_w$ is a decoder specific parametric function that returns the
unnormalized probability for the next target symbol being $w$.

\subsection{Learning}

Training this multi-way, multilingual model does not require multi-way parallel
corpora but only a set of bilingual corpora. For each bilingual pair, the
conditional log-probability of a ground-truth translation given a source
sentence is maximize by adjusting the relevant parameters following the gradient
of the log-probability.

\section{Translation Strategies}
\label{sec:strategies}

\subsection{One-to-One Translation}
\label{sec:one-to-one}

In the original paper by \newcite{firat2016multi}, only one translation strategy
was evaluated, that is, {\it one-to-one translation}. This one-to-one strategy
works on a source sentence given in one language by taking the encoder of that
source language, the decoder of a target language and the shared attention
mechanism. These three components are glued together as if they form a
single-pair neural translation model and translates the source sentence into a
target language. 

We however notice that this is not the only translation strategy available with
the multi-way, multilingual model. As we end up with multiple encoders, multiple
decoder and a shared attention mechanism, this model naturally enables us to
exploit a source sentence given in multiple languages, leading to a many-to-one
translation strategy which was proposed recently by \newcite{zoph2016multi} in
the context of neural machine translation. 

Unlike \cite{zoph2016multi}, the multi-way, multilingual model is not trained
with multi-way parallel corpora. This however does not necessarily imply that
the model cannot be used in this way. In the remainder of this section, we
propose two alternatives for doing multi-source translation with the multi-way,
multilingual model. 

\subsection{Many-to-One Translation}
\label{sec:N-to-one-exp}

In this section, we consider a case where a source sentence is given in two
languages, $X_1$ and $X_2$.  However, any of the approaches described below
applies to more than two source languages trivially.

In this multi-way, multilingual model, multi-source translation can be thought
of as averaging two separate translation paths. For instance, in the case of
Es+Fr to En, we want to combine Es$\to$En and Fr$\to$En so as to get a better
English translation. We notice that there are two points in the multi-way,
multilingual model where this averaging may happen. 

\paragraph{Early Average}
The first candidate is to averaging two translation paths when computing the
time-dependent context vector (see Eq.~\eqref{eq:cv}.) At each time $t$ in the
decoder, we compute a time-dependent context vector for each source language,
$\vc_t^1$ and $\vc_t^2$ respectively for the two source languages. In this early
averaging strategy, we simply take the average of these two context vectors:
\begin{align}
    \label{eq:early-avg}
    \vc_t = \frac{\vc_t^1 + \vc_t^2}{2}.
\end{align}
Similarly, we initialize the decoder's hidden state to be the average of the
initializers of the two encoders:
\begin{align}
    \label{eq:early-avg-init}
    \vz_0 = \frac{1}{2}\left( \phi_{\init}( \phi_{\init}^1 (\vh_{T_{x_1}}^1)) +
        \phi_{\init}( \phi_{\init}^2 (\vh_{T_{x_1}}^2))
    \right),
\end{align}
where $\phi_{\init}$ is the decoder's initializer (see Eq.~\eqref{eq:init}.)

\paragraph{Late Average}
Alternatively, we can average those two translation paths (e.g., Es$\to$En and
Fr$\to$En) at the output level. At each time $t$, each translation path computes
the distribution over the target vocabulary, i.e., $p(y_t=w|y_{<t}, X_1)$ and
$p(y_t=w|y_{<t}, X_2)$. We then average them to get the multi-source output
distribution:
\begin{align}
    \label{eq:late-avg}
    p(y_t=w|&y_{<t}, X_1, X_2) = \\
            & \frac{1}{2}(p(y_t=w|y_{<t}, X_1)+p(y_t=w|y_{<t})).
    \nonumber
\end{align}
An advantage of this late averaging strategy over the early averaging one is
that this can work even when those two translation paths were not from a single
multilingual model. They can be two separately single-pair models. In fact, if
$X_1$ and $X_2$ are same and the two translation paths are simply two different
models trained on the same language pair--direction, this is equivalent to
constructing an ensemble, which was found to greatly improve translation
quality~\cite{sutskever2014sequence}

\paragraph{Early+Late Average} 
The two strategies above can be further combined by late-averaging the output
distributions from the early averaged model and the late averaged one. We
empirically evaluate this early+late average strategy as well.

\section{Experiments: Translation Strategies and Multi-Source Translation}
\label{sec:strategies-exp}

Before continuing on with zero-resource machine translation, we first evaluate
the translation strategies described in the previous section on multi-source
translation, as these translation strategies form a basic foundation on which
we extend the multi-way, multilingual model for zero-resource machine
translation.

\subsection{Settings}
\label{sec:settings}

When evaluating the multi-source translation strategies, we use English, Spanish
and French, and focus on a scenario where only En-Es and En-Fr parallel corpora
are available. 

\subsubsection{Corpora}
\label{sec:corpora}

\paragraph{En-Es} 
We combine the following corpora to form 34.71m parallel Es-En sentence pairs:
UN (8.8m), Europarl-v7 (1.8m), news-commentary-v7 (150k), LDC2011T07-T12 (2.9m)
and internal technical-domain data (21.7m). 

\paragraph{En-Fr}
We combine the following corpora to form 65.77m parallel En-Fr sentence pairs:
UN (9.7m), Europarl-v7 (1.9m), news-commentary-v7 (1.2m), LDC2011T07-T10 (1.6m),
ReutersUN (4.5m), internal technical-domain data (23.5m) and Gigaword R2
(20.66m). 

\paragraph{Evaluation Sets}
We use newstest-2012 and newstest-2013 from WMT as development and test sets,
respectively.

\paragraph{Monolingual Corpora}
We do not use any additional monolingual corpus.

\paragraph{Preprocessing}
All the sentences are tokenized using the tokenizer script from
Moses~\cite{koehn2007moses}. We then replace special tokens, such as numbers,
dates and URL's with predefined markers, which will be replaced back with the
original tokens {\it after} decoding. After using byte pair encoding (BPE,
\cite{sennrich2015neural}) to get subword symbols, we end up with 37k, 43k
and 45k unique tokens for English, Spanish and French, respectively. For
training, we only use sentence pairs in which both sentences are only up to 50
symbols long.

See Table~\ref{table:stats} for the detailed statistics.

\begin{table}
\centering
\small

\begin{minipage}{0.7\linewidth}
    \begin{tabular}{l|ccc}
    \# Sents & Train & Dev$^\dagger$ & Test$^\ddagger$ \\
        \toprule
    En-Es             & 34.71m                 & 3003                 & 3000                  \\
    En-Fr             & 65.77m                 & 3003                 & 3000                  \\ 
        \midrule
    En-Es-Fr          & 11.32m                 & 3003                 & 3000                 
    \end{tabular}
\end{minipage}
\begin{minipage}{0.27\linewidth}
\caption{Data statistics.
$\dagger$: newstest-2012. $\ddagger$: newstest-2013}
\label{table:stats}
\end{minipage}

\vspace{-4mm}
\end{table}

\subsection{Models and Training}
\label{sec:models}

We start from the code made publicly available as a part of
\cite{firat2016multi}.\footnote{
    \url{https://github.com/nyu-dl/dl4mt-multi}
} We made two changes to the original code. First, we replaced the decoder with
the conditional gated recurrent network with the attention mechanism as outlines
in \cite{firat2016}. Second, we feed a binary indicator vector of which
encoder(s) the source sentence was processed by to the output layer of each
decoder ($g^m_w$ in Eq.~\eqref{eq:output}). Each dimension of the indicator vector
corresponds to one source language, and in the case of multi-source translation,
there may be more than one dimensions set to 1.

We train the following models: four single-pair models (Es$\leftrightarrow$En
and Fr$\leftrightarrow$En) and one multi-way, multilingual model
(Es,Fr,En$\leftrightarrow$Es,Fr,En). As proposed by \newcite{firat2016multi},
we share one attention mechanism for the latter case.

\paragraph{Training} 

We closely follow the setup from \cite{firat2016multi}. Each symbol is
represented as a 620-dimensional vector. Any recurrent layer, be it in the
encoder or decoder, consists of 1000 gated recurrent units (GRU,
\cite{cho2014learning}), and the attention mechanism has a hidden layer of 1200
$\tanh$ units ($f_{\text{score}}$ in Eq.~\eqref{eq:att}). We use Adam~\cite{kingma2014adam} to
train a model, and the gradient at each update is computed using a minibatch of
at most 80 sentence pairs. The gradient is clipped to have the norm of at most
1~\cite{pascanu2012difficulty}. We early-stop any training using the T-B
score on a development set.\footnote{
    T-B score is defined as $\frac{\text{TER}-\text{BLEU}}{2}$ which we found
    to be more stable than either TER or BLEU alone for the purpose of
    early-stopping~\cite{zhao2009simplex}.
}

\begin{table}
    \centering
    \small 
    \begin{tabular}{c | c c || c c | c c}
        & & & \multicolumn{2}{c|}{Multi} & \multicolumn{2}{c}{Single} \\
        & Src & Trgt & Dev & Test & Dev & Test \\
        \toprule
        (a) & Es & En & 30.73 & 28.32 & 29.74 & 27.48 \\
        (b) & Fr & En & 26.93 & 27.93 & 26.00 & 27.21 \\
        \midrule
        (c) & En & Es & 30.63 & 28.41 & 31.31 & 28.90 \\
        (d) & En & Fr & 22.68 & 23.41 & 22.80 & 24.05 \\
    \end{tabular}
    \caption{One-to-one translation qualities using the multi-way, multilingual
    model and four separate single-pair models.}
    \label{table:one-to-one}
    \vspace{-4mm}
\end{table}

\subsection{One-to-One Translation}

We first confirm that the multi-way, multilingual translation model indeed works
as well as single-pair models on the translation paths that were considered
during training, which was the major claim in \cite{firat2016multi}. In
Table~\ref{table:one-to-one}, we present the results on four language
pair-directions (Es$\leftrightarrow$En and Fr$\leftrightarrow$En).

It is clear that the multi-way, multilingual model indeed performs comparably on
all the four cases with less parameters (due to the shared attention mechanism.)
As observed earlier in \cite{firat2016multi}, we also see that the multilingual
model performs better when a target language is English.

\begin{table}
    \centering
    \begin{tabular}{ c | c || c c | c c}
        \multicolumn{2}{c||}{} & \multicolumn{2}{c|}{Multi} & \multicolumn{2}{c}{Single} \\
        \multicolumn{2}{c||}{}  & Dev & Test & Dev & Test \\
        \toprule
        (a) & Early & 31.89 & 31.35 & -- & -- \\
        (b) & Late & 32.04 & 31.57 & 32.00 & 31.46 \\
        (c) & E+L & 32.61 & 31.88 & -- & -- \\
    \end{tabular}
    \caption{Many-to-one quality (Es+Fr$\to$En) using three translation
    strategies. Compared to Table~\ref{table:one-to-one}~(a--b) we observe a
significant improvement (up to 3+ BLEU), although the model was never trained in
these many-to-one settings. The second column shows the quality by the ensemble
of two separate single-pair models.}
    \label{table:N-to-one}
    \vspace{-4mm}
\end{table}

\subsection{Many-to-One Translation}
\label{sec:N-to-one}

We consider translating from a pair of source sentences in Spanish (Es) and
French (Fr) to English (En). It is important to note that the multilingual model
was {\em not} trained with any multi-way parallel corpus. Despite this, we
observe that the early averaging strategy improves the translation quality
(measured in BLEU) by 3 points in the case of the test set (compare
Table~\ref{table:one-to-one}~(a--b) and Table~\ref{table:N-to-one}~(a).) We
conjecture that this happens as training the multilingual model has implicitly
encouraged the model to find a {\it common context vector} space across multiple
source languages.

The late averaging strategy however outperforms the early averaging in both
cases of multilingual model and a pair of single-pair models (see
Table~\ref{table:N-to-one}~(b)) albeit marginally. The best quality was observed
when the early and late averaging strategies were combined at the output level,
achieving up to +3.5 BLEU (compare Table~\ref{table:one-to-one}~(a) and
Table~\ref{table:N-to-one}~(c).)

We emphasize again that there was no multi-way parallel corpus consisting of
Spanish, French and English during training. The result presented in this
section shows that the multi-way, multilingual model can exploit multiple
sources effectively without requiring any multi-way parallel corpus, and we will
rely on this property together with the proposed many-to-one translation
strategies in the later sections where we propose and investigate zero-resource
translation.

\section{Zero-Resource Translation Strategies}

The network architecture of multi-way, multilingual model suggests the potential
for translating between two languages {\it without} any direct parallel corpus
available. In the setting considered in this paper (see
Sec.~\ref{sec:settings},) these translation paths correspond to
Es$\leftrightarrow$Fr, as only parallel corpora used for training were
Es$\leftrightarrow$En and Fr$\leftrightarrow$En. 

The most naive approach for translating along a zero-resource path is to simply
treat it as any other path that was included as a part of training. This
corresponds to the one-to-one strategy from Sec.~\ref{sec:one-to-one}. In our
experiments, it however turned out that this naive approach does not work at
all, as can be seen in Table~\ref{table:zr-trans}~(a).

In this section, we investigate this potential of zero-resource translation with
the multi-way, multilingual model in depth. More specifically, we propose a
number of approaches that enable zero-resource translation without requiring
any additional bilingual corpus.

\subsection{Pivot-based Translation}

The first set of approaches exploits the fact that the target zero-resource
translation path can be decomposed into a sequence of high-resource translation
paths~\cite{wu2007pivot,utiyama2007comparison}. For instance, in our case,
Es$\to$Fr can be decomposed into a sequence of Es$\to$En and En$\to$Fr. In other
words, we translate a source sentence (Es) into a pivot language (En) and then
translate the English translation into a target language (Fr). 

\paragraph{One-to-One Translation}

The most basic approach here is to perform each translation path in the
decomposed sequence independently from each other. This one-to-one approach
introduces only a minimal computational complexity (the multiplicative factor of
two.) We can further improve this one-to-one pivot-based translation by
maintaining a set of $k$-best translations from the first stage (Es$\to$En), but
this increase the overall computational complexity by the factor of $k$, making
it impractical in practice. We therefore focus only on the former approach of
keeping the best pivot translation in this paper.

\paragraph{Many-to-One Translation}

With the multi-way, multilingual model considered in this paper, we can extend
the naive one-to-one pivot-based strategy by replacing the second stage
(En$\to$Fr) to be many-to-one translation from Sec.~\ref{sec:N-to-one} using
both the original source language and the pivot language as a pair of source
languages. We first translate the source sentence (Es) into English, and use
both the original source sentence and the English translation (Es+En) to
translate into the final target language (Fr). 

Both approaches described and proposed above do not require any additional
action on an already-trained multilingual model. They are simply different
translation strategies specifically aimed at zero-resource translation.

\subsection{Finetuning with Pseudo Parallel Corpus}
\label{sec:finetuning}

The failure of the naive zero-resource translation earlier (see
Table~\ref{table:zr-trans}~(a)) suggests that the context vectors returned by
the encoder are not compatible with the decoder, when the combination was not
included during training. The good translation qualities of the translation
paths included in training however imply that the representations learned by the
encoders and decoders are good. Based on these two observations, we conjecture
that all that is needed for a zero-resource translation path is a simple
adjustment that makes the context vectors from the encoder to be compatible with
the target decoder. Thus, we propose to adjust this zero-resource translation
path however without any additional parallel corpus. 

First, we generate a small set of {\it pseudo bilingual pairs} of sentences for
the zero-resource language pair (Es$\to$Fr) in interest. We randomly select $N$
sentences pairs from a parallel corpus between the target language (Fr) and a
pivot language (En) and translate the pivot side (En) into the source language
(Es). Then, the pivot side is discarded, and we construct a {\it pseudo}
parallel corpus consisting of sentence pairs of the source and target languages
(Es-Fr). 

We make a copy of the existing attention mechanism, to which we refer as {\it
target-specific attention mechanism}.  We then finetune only this
target-specific attention mechanism while keeping all the other parameters of
the encoder and decoder intact, using the generated pseudo parallel corpus. 
We do not update any other parameters in the encoder and decoder, because they
are already well-trained (evidenced by high translation qualities in
Table~\ref{table:one-to-one}) and we want to avoid disrupting the well-captured
structures underlying each language. 

Once the model has been finetuned with the pseudo parallel corpus, we can use
any of the translation strategies described earlier in Sec.~\ref{sec:strategies}
for the finetuned zero-resource translation path. We expect a similar gain by
using many-to-one translation, which we empirically confirm in the next section.

\section{Experiments: \\~~~~~~Zero-Resource Translation}

\subsection{Without Finetuning}
\label{sec:no-finetuning}

\subsubsection{Settings}

We use the same multi-way, multilingual model trained earlier in
Sec.~\ref{sec:models} to evaluate the zero-resource translation strategies. We
emphasize here that this model was trained only using Es-En and Fr-En {\it
bilingual} parallel corpora without any Es-Fr parallel corpus.

\begin{table}[t]
    \centering
    \small
    \begin{tabular}{c || c | c || c c}
        & Pivot & Many-to-1 & Dev & Test \\
        \toprule
        (a) & & & $< 1$ & $< 1$ \\
        \midrule
        (b) & $\surd$ & & 20.64 & 20.4 \\
        \midrule
        (c) & $\surd$ & Early & 9.24 & 10.42 \\
        (d) & $\surd$ & Late & 18.22 & 19.14 \\
        (e) & $\surd$ & E+L & 13.29 & 14.56 \\
    \end{tabular}
    \caption{Zero-resource translation from Spanish (Es) to French (Fr) {\it
        without} finetuning. When
    pivot is $\surd$, English is used as a pivot language.}
    \label{table:zr-trans}
    \vspace{-4mm}
\end{table}

We evaluate the proposed approaches to zero-resource translation with the same
multi-way, multilingual model from Sec.~\ref{sec:settings}. We specifically
select the path from Spanish to French (Es$\to$Fr) as a target zero-resource
translation path. 

\subsubsection{Result and Analysis}

As mentioned earlier, we observed that the multi-way, multilingual model {\it
cannot} directly translate between two languages when the translation path
between those two languages was not included in training
(Table~\ref{table:zr-trans}~(a).) On the other hand, the model was able to
translate decently with the pivot-based one-to-one translation strategy, as can
be see in Table~\ref{table:zr-trans}~(b). Unsurprisingly, all the many-to-one
strategies resulted in worse translation quality, which is due to the inclusion
of the useless translation path (direct path between the zero-resource pair,
Es-Fr.) These results clearly indicate that the multi-way, multilingual model
trained with only bilingual parallel corpora is not capable of direct
zero-resource translation {\it as it is}.

\begin{table*}[t]
    \centering
    \begin{tabular}{c | c c c || c c c c | c c c c}
        & & & & \multicolumn{4}{c|}{Pseudo Parallel Corpus} &
        \multicolumn{4}{c}{True Parallel Corpus} \\
        & Pivot & Many-to-1 & &  1k & 10k & 100k & 1m & 1k & 10k & 100k & 1m \\
        \toprule
        \multirow{2}{*}{(a)} & \multicolumn{2}{c}{\multirow{2}{*}{Single-Pair Models}} & 
        Dev  & -- & -- & -- & -- & -- & -- & 11.25 & 21.32 \\ 
             & \multicolumn{2}{c}{}&
        Test & -- & -- & -- & -- & -- & -- & 10.43 & 20.35 \\ 
        \midrule
        (b) & $\surd$ & \multicolumn{2}{c||}{No Finetuning} & 
        \multicolumn{4}{c|}{Dev: 20.64, Test: 20.4} &
        \multicolumn{4}{c}{--}\\
        \midrule
        \multirow{2}{*}{(c)} & \multirow{2}{*}{} & \multirow{2}{*}{} & 
        Dev  & 0.28 & 10.16 & 15.61 & 17.59 & 0.1 & 8.45 & 16.2 & 20.59 \\ 
         & & & 
        Test & 0.47 & 10.14 & 15.41 & 17.61 & 0.12 & 8.18 & 15.8 & 19.97 \\ 
        \midrule
        \multirow{2}{*}{(d)} & \multirow{2}{*}{$\surd$} & \multirow{2}{*}{Early} & 
        Dev  & 19.42 & 21.08 & 21.7  & 21.81 & 8.89 & 16.89  & 20.77  & 22.08  \\ 
         & & & 
        Test & 19.43 & 20.72 & 21.23 & 21.46 & 9.77 & 16.61  & 20.40  & 21.7  \\ 
        \midrule
        \multirow{2}{*}{(e)} & \multirow{2}{*}{$\surd$} & Early+ & 
        Dev  & 20.89 & 20.93 & 21.35 & 21.33 & 14.86 & 18.28 & 20.31 & 21.33 \\ 
         & & Late & 
        Test & 20.5 & 20.71 & 21.06  & 21.19 & 15.42 & 17.95 & 20.16 & 20.9 \\ 
    \end{tabular}
    \caption{Zero-resource translation from Spanish (Es) to French (Fr) {\it
        with} finetuning. When
    pivot is $\surd$, English is used as a pivot language. Row (a) is from
Table~\ref{table:zr-trans}~(b). 
}
    \label{table:zr-trans-finetune}
    \vspace{-4mm}
\end{table*}

\subsection{Finetuning with a Pseudo Parallel Corpus}

\subsubsection{Settings}

The proposed finetuning strategy raises a number of questions. First, it is
unclear how many pseudo sentence pairs are needed to achieve a decent
translation quality. Because the purpose of this finetuning stage is simply to
adjust the shared attention mechanism so that it can properly bridge from the
source-side encoder to the target-side decoder, we expect it to work with only a
small amount of pseudo pairs. We validate this by creating pseudo corpora of
different sizes--1k, 10k, 100k and 1m. 

Second, we want to know how detrimental it is to use the generated pseudo
sentence pairs compared to using true sentence pairs between the target language
pair.  In order to answer this question, we compiled a true multi-way parallel
corpus by combining the subsets of UN (7.8m), Europral-v7 (1.8m),
OpenSubtitles-2013 (1m), news-commentary-v7 (174k), LDC2011T07 (335k) and
news-crawl (310k), and use it to finetune the model.\footnote{
    See the last row of Table~\ref{table:stats}.
}
This allows us to evaluate the effect of the pseudo and true parallel corpora on
finetuning for zero-resource translation.

Lastly, we train single-pair models translating directly from Spanish to French
by using the true parallel corpora. These models work as a baseline against
which we compare the multi-way, multilingual models.

\paragraph{Training}
Unlike the usual training procedure described in Sec.~\ref{sec:models}, we
compute the gradient for each update using 60 sentence pairs only, when
finetuning the model with the multi-way parallel corpus (either pseudo or true.) 

\subsubsection{Result and Analysis}

Table~\ref{table:zr-trans-finetune} summarizes all the result. The most
important observation is that the proposed finetuning strategy with {\it
pseudo}-parallel sentence pairs outperforms the pivot-based approach (using the
early averaging strategy from Sec.~\ref{sec:N-to-one}) even when we used only
1,000 such pairs (compare (b) and (d).) As we increase the size of the
pseudo-parallel corpus, we observe a clear improvement. Furthermore, these
models perform comparably to or better than the single-pair model trained with
1M {\it true} parallel sentence pairs, {\it although they never saw a single
true bilingual sentence pair} of Spanish and French (compare (a) and (d).) Even
when we trained a single-pair model with 11m true parallel pairs, the model
could not match the multilingual model finetuned with 1m true parallel pairs by
achieving the translation quality of 24.26 BLEU on the test set.

Another interesting finding is that it is only beneficial to use true parallel
pairs for finetuning the multi-way, mulitilingual models when there are enough
of them (1m or more). When there are only a small number of true parallel
sentence pairs, we even found using pseudo pairs to be more beneficial than true
ones. This effective as more apparent, when the direct one-to-one translation of
the zero-resource pair was considered (see (c) in
Table~\ref{table:zr-trans-finetune}.) This applies that the misalignment between
the encoder and decoder can be largely fixed by using pseudo-parallel pairs
only, and we conjecture that it is easier to learn from pseudo-parallel pairs
as they better reflect the inductive bias of the trained model.  When there is a
large amount of true parallel sentence pairs available, however, our results
indicate that it is better to exploit them.

Unlike we observed with the multi-source translation in
Sec.~\ref{sec:N-to-one-exp}, we were not able to see any improvement by further
averaging the early-averaged and late-average decoding schemes (compare (d) and
(e).) This may be explained by the fact that the context vectors computed when
creating a pseudo source (e.g., En from Es when Es$\to$Fr) already contains all
the information about the pseudo source. It is simply enough to take those
context vectors into account via the early averaging scheme.

These results clearly indicate and verify the potential of the multi-way,
multilingual neural translation model in performing zero-resource machine
translation. More specifically, it has been shown that the translation quality
can be improved even without any direct parallel corpus available, and if there
is a small amount of direct parallel pairs available, the quality may improve
even further. 

\section{Conclusion: \\
~~~~~~Implications and Limitations}

\paragraph{Implications}

There are two main results in this paper. First, we showed that the multi-way,
multilingual neural translation model by \newcite{firat2016multi} is able to
exploit common, underlying structures across many languages in order to better
translate when a source sentence is given in multiple languages. This confirms
the usefulness of positive language transfer, which has been believed to be an
important factor in human language learning~\cite{odlin1989,ringbom2007cross},
in machine translation. Furthermore, our result significantly expands the
applicability of multi-source translation~\cite{zoph2016multi}, as it does not
assume the availability of multi-way parallel corpora for training. 

Second, the experiments on zero-resource translation revealed that it is not
necessary to have a direct parallel corpus, or deep linguistic knowledge,
between two languages in order to build a machine translation system.
Importantly we observed that the proposed approach of zero-resource translation
is better both in terms of translation quality and data efficiency than a more
traditional pivot-based translation~\cite{wu2007pivot,utiyama2007comparison}.
Considering that this is the first attempt at such zero-resource, or extremely
low-resource, translation using neural machine translation, we expect a large
progress in near future. 

\paragraph{Limitations}

Despite the promising empirical results presented in this paper, there are a
number of shortcomings that needs to addressed in follow-up research. First, our
experiments have been done only with three European languages--Spanish, French
and English. More investigation with a diverse set of languages needs to be done
in order to make a more solid conclusion, such as was done in
\cite{firat2016multi,chung2016character}. Furthermore, the effect of varying
sizes of available parallel corpora on the performance of zero-resource
translation must be studied more in the future.  

Second, although the proposed many-to-one translation is indeed generally
applicable to any number of source languages, we have only tested a source
sentence in two languages. We expect even higher improvement with more
languages, but it must be tested thoroughly in the future.

Lastly, the proposed finetuning strategy requires the model to have an
additional set of parameters relevant to the attention mechanism for a target,
zero-resource pair. This implies that the number of parameters may grow linearly
with respect to the number of target language pairs. We expect future research
to address this issue by, for instance, mixing in the parallel corpora of
high-resource language pairs during finetuning as well.

\section*{Acknowledgments}
OF thanks Georgiana Dinu and Iulian Vlad Serban for insightful discussions. KC
thanks the support by Facebook, Google (Google Faculty Award 2016) and NVidia
(GPU Center of Excellence 2015-2016). 

\vspace{-15px}
\bibliography{zrnmt}
\bibliographystyle{emnlp2016}

\end{document}